\documentclass[10pt,letterpaper]{article}
\usepackage[top=0.85in,left=2.75in,footskip=0.75in]{geometry}
\usepackage[utf8]{inputenc}

\usepackage{amssymb}
\usepackage{amsfonts}
\usepackage{rotating,listings}
\usepackage{amsmath}

\usepackage{changepage}

\usepackage{textcomp,marvosym}

\usepackage{cite}

\usepackage{nameref,hyperref}

\usepackage{comment} 
\usepackage{cite}
\usepackage{color}
\usepackage{graphicx}

\usepackage[right]{lineno}

\usepackage{microtype}
\DisableLigatures[f]{encoding = *, family = * }

\usepackage[table]{xcolor}

\usepackage{array}

\newcolumntype{+}{!{\vrule width 2pt}}

\newlength\savedwidth

\raggedright
\usepackage[labelfont=bf,labelsep=period,justification=raggedright]{caption}

\makeatletter
\renewcommand{\@biblabel}[1]{\quad#1.}
\makeatother

\usepackage{subscript}

\usepackage{lastpage,fancyhdr,graphicx}
\usepackage{epstopdf}
\fancyheadoffset[L]{2.25in}
\fancyfootoffset[L]{2.25in}
\usepackage{color}
\usepackage{url}

\newcommand{\be}{\begin{equation}}
\newcommand{\ee}{\end{equation}}

\newcommand{\bea}{\begin{eqnarray}}
\newcommand{\eea}{\end{eqnarray}}

\usepackage{longtable}
\usepackage{arydshln}

\begin{document}

\vspace*{0.2in}


\begin{flushleft}
{\Large
\textbf{Mapping the Americanization of English in Space and Time}
}
\newline
\\
Bruno Gon\c calves $^{1}$, 
Luc\'ia Loureiro-Porto$^{2}$,
Jos\'e J. Ramasco$^{3}$,
David S\'anchez$^{3,*}$
\\
\bigskip
\bf{1} Center for Data Science, New York University, 60 5$^{th}$ Avenue, New York, NY 10012, USA\\
\bf{2} Departament de Filologia Espanyola, Moderna i Cl\`assica, Universitat de les Illes Balears (UIB), 07122 Palma de Mallorca, Spain\\
\bf{3} Institute for Cross-Disciplinary Physics and Complex Systems IFISC (CSIC-UIB), 07122 Palma de Mallorca, Spain\\
\bigskip

$*$ E-mail: david.sanchez@uib.es\\
\end{flushleft}

\section*{Abstract}

As global political preeminence gradually shifted from the United Kingdom to the United States, so did the capacity to culturally influence the rest of the world. In this work, we analyze how the world-wide varieties of written English are evolving. We study both the spatial and temporal variations of vocabulary and spelling of English using a large corpus of geolocated tweets and the Google Books datasets corresponding to books published in the US and the UK.
The advantage of our approach is that we can address both standard written language (Google Books)
and the more colloquial forms of microblogging messages (Twitter).
We find that American English is the dominant form of English outside the UK and that its influence is felt even within the UK borders. Finally, we analyze how this trend has evolved over time and the impact that some cultural events have had in shaping it.


\section*{Introduction}
With roots dating as far back as Cabot's explorations in the 15th century and the $1584$ establishment of the ill-fated Roanoke colony in the New World, the British empire was one of the largest empires in Human History. At its zenith, it extended from North America to Asia, Africa and Australia deserving the moniker ``the empire on which the sun never sets''. However, as history has shown countless times, empires rise and fall due to a complex set of internal and external forces. In the case of the British empire, its  preeminence faded as the United States of America --one of its first colonies-- took over the dominant role in the global arena. 

As an empire spreads so does the language of its ruling class
because the presence of a prestigious linguistic variety plays an important role in language change~\cite{labov63,fishman67}.
Thanks to both its global extension, late demise, and the rise of the US as a global actor, the English language enjoys an undisputed role as the global \emph{lingua franca} serving as the default language of science, commerce and diplomacy~\cite{crystal03-1, jenkins13-1} (see Fig~\ref{tweets}). Given such an extended presence, it is only natural that English would absorb words, expressions and other features of local indigenous languages resulting in dozens of dialects and topolects (language forms typical of a specific area) such as ``Singlish'' (Singapore), ``Hinglish'' (India), Kenyan English~\cite{mesthrie08-1}, and, most importantly, American English~\cite{grieve16-1} a variety that includes within itself several other dialects~\cite{pederson01-1,wolfram16-1}.

The transfer of political, economical and cultural power from Great Britain to the United States has progressed gradually over the course of more than half a century, with World War II being the final stepping stone in the establishment of American supremacy. The cultural rise of the United States also implied the exportation of their specific form of English resulting in a change of how English is written and spoken around the world. In fact, the ``Americanization'' of (global) English is one of the main processes of language change in contemporary English~\cite{leech09-1}.
Although it is found to work along with other processes such as colloquialization and informalization~\cite{baker17}, the spread of American features all over the globe is generally assumed to be result of the American ‘leadership’ in change~\cite{leech09-1}.

As an example, if we focus on spelling, some the original differences between British and American English orthography (most of which are the result of Webster's reform~\cite{algeo01-1}) are somehow blurred and, for instance, the tendency for verbs and nouns to end in \textit{-ize} and \textit{-ization} in America is now common on both sides of the Atlantic~\cite{gramley03-1}. Likewise, a tendency for Postcolonial varieties of English in South-East Asia to prefer American spelling over the British one has been observed, at least, for Nigerian English~\cite{awonusi94-1}, Singapore and Trinidad and Tobago~\cite{hansel13-1}, regarding spelling and lexis, for Indian English~\cite{davydova16-1} and the Bahamas~\cite{hackert15-1}, regarding syntax, and for Hong Kong~\cite{edwards16-1}, regarding phonology. In addition, a growing tendency for Americanization has been observed for Philippine English, which, despite being rooted in American English, has experienced a rise in the frequency of American forms~\cite{fuchs17-1}. Although this Americanization is found in different registers, web genres have been highlighted as a text-type where American forms are preferred ~\cite{mukherjee15-1}. Electronic communication has indeed been considered to play a role in linguistic uniformity\cite{venezky01-1}. It is in this sense that this paper will make a contribution to the study of the Americanization of English, since a corpus of $213,086,831$ geolocated tweets will be used to study the spread of American English spelling and vocabulary around the globe, including regions where English is used as a first, second and foreign language.

The study of diatopic variation using Twitter datasets is a relatively new subject~\cite{nguyen15-1}.
The use of geotagged microblogging data~\cite{melo16-1}
allows the quantitative examination of linguistic patterns on a worldwide scale, in automatic fashion and within conversational situations. The global extension and the real time availability of the data constitute major methodological advantages over more traditional approaches like surveys and interviews ~\cite{chambers98-1}.
Importantly, the resulting corpora are publicly available~\cite{malmasi}, although due to their nature
most of the literature has been concerned with lexical variation (for an exception that addresses semantic and syntactic variation,
see Ref.~\cite{kulkarni16-1}). Thus, different variables can be mapped after carefully removing
lexical ambiguities~\cite{russ12-1}. A Bayesian approach shows good agreement between baseline queries and survey responses~\cite{doyle14-1}. Machine learning techniques applied to Twitter corpora reveal the existence of superdialects~\cite{goncalves14-1,goncalves16-1}, which
can be further analyzed with dialectometric techniques~\cite{donoso}. Linguistic evolution in social media appears to be strongly connected to demographics~\cite{eisenstein14-1}. Age and gender issues can be additionally introduced in the analysis~\cite{pavalanathan15-1}.
Moreover, an investigation of lexical alternations unveils hierarchical dialect regions in the United States~\cite{huang16-1}.
Twitter can be also employed in the study of specific varieties departing from the standard form~\cite{blodgett16-1}.
However, online social media are more suitable for a synchronic approximation to language variation.
If one aims at understanding the diachronic evolution of language, we need a corpus well established over time.
This is available with the Google Books database~\cite{michel11-1}, which has already been used for the analysis of relative frequencies
that characterize word fluxes~\cite{pedersen12-1,pechenik15-1} or the applicability of Zip's and Heaps's law with different scaling
regimes~\cite{gerlach13-1}. Here, we will complement our Twitter study of the Americanization of English with an analysis
of the dynamic process that is taking place since $1800$.

In this paper we analyze how English is used around the world, in informal contexts, using a large scale Twitter dataset. Due to the written nature of our corpus we consider in detail both how vocabulary and spelling of common words varies from place to place in order to understand how American cultural influence is spreading around the world. We complement this synchronic analysis with a diachronic view of how the prevalence of British and American vocabulary and spelling have evolved over time in British and American publications using the Google Books dataset.

\section*{Methods}

\subsection*{Datasets}

The goal of this manuscript is to analyze how English is used across both time and space. We study the geographical variation of English by using the Twitter Decahose from which we collect\cite{mocanu13-1} all tweets written in English between May $10, 2010$ and Feb $28, 2016$ that contain geolocation information (see supporting information below). The language is detected using Chromium Compact Language Detection library as in Ref.~\cite{mocanu13-1}. One might ask whether those tweets arising from outside the English-speaking world are from native-English speakers residing in or visiting those countries.
In fact, it has been shown that the vast majority of the tweets in a given location arises from speakers residing in that country~\cite{native1,native2}.
Therefore, our approach is reliable to a very good extent
but has the limitations common to geographical studies based on Twitter datasets,
as illustrated, e.g., in Ref.~\cite{leetaru13}.

The temporal evolution of English is analyzed using the Google Books dataset\cite{michel11-1} of books published by both British and American publishers (see supporting information below). The dataset contains the number of times individual words were used in books scanned by Google and dating back to the 15th century. However, due to the poor statistics in earlier periods, we restrict our analysis to the period between $1800$ and $2010$. Importantly, for a given year in our dataset each of the two corpora (British and American) include at least several million word instances, ranging from a minimum 18.4 million for US in 1800 (98.5 million in the UK) to a maximum of 8.2 billion for the US in 2000 (2.2 billion in the UK). To the best of our knowledge, these are the largest corpora of this kind ever gathered.
Admittedly, the Google Books dataset has its own shortcomings (prolific authors, overrepresentation of scientific texts, etc.~\cite{pechenik15}) but these
are expected to affect both corpora equally.

Our two main data sources are different in nature: Twitter contains more colloquial expressions, while the language recorded in the books is more formal.
As a result, these two sources, in combination, can provide a useful perspective on the spatio-temporal patterns developed or developing in English.
Yet, in our study we do not distinguish between the two types. Rather, our objective is to show that, despite the fact that both corpora have different register features, the English Americanization is evident in the two of them.

\begin{longtable}{p{.50\textwidth} | p{.50\textwidth}}
\hline
\hline
British&American\\
\hline
railway& railroad\\ 
\hdashline
MA dissertation& MA thesis\\ 
\hdashline
doctoral thesis& doctoral dissertation\\ 
\hdashline
draughts& checkers\\ 
\hdashline
abseil& rappel\\ 
\hdashline
antenatal& prenatal\\ 
\hdashline
anticlockwise& counterclockwise\\ 
\hdashline
aubergine& eggplant\\ 
\hdashline
barrister, solicitor& attorney\\ 
\hdashline
biscuit& cookie\\ 
\hdashline
car park& parking lot\\ 
\hdashline
caster sugar, icing sugar& confectioner's sugar, powdered sugar\\ 
\hdashline
corn flour& corn starch\\ 
\hdashline
cupboard & closet\\ 
\hdashline
demister& defroster\\ 
\hdashline
drawing pin& thumbtack\\ 
\hdashline
Father Christmas& Santa Claus\\ 
\hdashline
handbrake, hand brake& emergency brake\\ 
\hdashline
hire purchase& installment plan\\ 
\hdashline
inside leg& inseam\\ 
\hdashline
mobile phone& cell phone\\ 
\hdashline
motorway& expressway, freeway\\ 
\hdashline
nappy& diaper\\ 
\hdashline
notice board& bulletin board\\ 
\hdashline
number plate& license plate\\ 
\hdashline
plasterboard& wallboard\\ 
\hdashline
polystyrene& styrofoam\\ 
\hdashline
porridge& oatmeal\\ 
\hdashline
perspex& plexiglass\\ 
\hdashline
pushchair& stroller\\ 
\hdashline
rubbish& garbage\\ 
\hdashline
skirting board& baseboard\\ 
\hdashline
spring onion& green onion\\ 
\hdashline
sticky tape& scotch tape\\ 
\hdashline
sweets& candy\\ 
\hdashline
torch& flashlight\\ 
\hdashline
tracksuit& sweatsuit\\ 
\hdashline
trousers& pants\\ 
\hdashline
valuer& appraiser\\ 
\hdashline
wellington boots, wellingtons& rubbers, rubber boots, rain boots\\ 
\hdashline
windscreen& windshield\\ 
\hdashline
lorry& truck\\ 
\hdashline
chemist's& drug store\\ 
\hdashline
elastic band& rubber band\\ 
\hdashline
estate agent& realtor\\ 
\hdashline
cot& crib\\ 
\hdashline
off-licence& liquor store\\ 
\hdashline
crayfish& crawfish\\ 
\hline
\hline
\caption{Word list comprising British and American vocabulary variants.} 
\label{tab:vocabulary}
\end{longtable}

\begin{longtable}{p{.50\textwidth} | p{.50\textwidth}}
\hline
\hline
British&American\\
\hline
skilful& skillful\\ 
\hdashline
wilful& willful\\ 
\hdashline
fulfil, fulfils& fulfill, fulfills\\ 
\hdashline
instil, instils& instill, instills\\ 
\hdashline
appal, appals& appall, appalls\\ 
\hdashline
flavour& flavor\\ 
\hdashline
mould& mold\\ 
\hdashline
moult& molt\\ 
\hdashline
smoulder& smolder\\ 
\hdashline
moustache& mustache\\ 
\hdashline
centre& center\\ 
\hdashline
metre& meter\\ 
\hdashline
theatre& theater\\ 
\hdashline
analyse& analyze\\ 
\hdashline
paralyse& paralyze\\ 
\hdashline
defence& defense\\ 
\hdashline
offence& offense\\ 
\hdashline
pretence& pretense\\ 
\hdashline
revelling, revelled& reveling, reveled\\ 
\hdashline
travelled, travelling& traveled, traveling\\ 
\hdashline
travelle& traveler\\ 
\hdashline
marvellous& marvelous\\ 
\hdashline
plough& plow\\ 
\hdashline
aluminium& aluminum\\ 
\hdashline
jewellery& jewelry\\ 
\hdashline
pyjamas& pajamas\\ 
\hdashline
whisky& whiskey\\ 
\hdashline
neighbour& neighbor\\ 
\hdashline
honour& honor\\ 
\hdashline
colour& color\\ 
\hdashline
behaviour& behavior\\ 
\hdashline
labour& labor\\ 
\hdashline
humour& humor\\ 
\hdashline
favour& favor\\ 
\hdashline
harbour& harbor\\ 
\hdashline
tumour& tumor\\ 
\hdashline
vigour& vigor\\ 
\hdashline
rumour& rumor\\ 
\hdashline
rigour& rigor\\ 
\hdashline
demeanour& demeanor\\ 
\hdashline
clamour& clamor\\ 
\hdashline
odour& odor\\ 
\hdashline
armour& armor\\ 
\hdashline
endeavour& endeavor\\ 
\hdashline
parlour& parlor\\ 
\hdashline
vapour& vapor\\ 
\hdashline
saviour& savior\\ 
\hdashline
splendour& splendor\\ 
\hdashline
fervour& fervor\\ 
\hdashline
savour& savor\\ 
\hdashline
valour& valor\\ 
\hdashline
candour& candor\\ 
\hdashline
ardour& ardor\\ 
\hdashline
rancour& rancor\\ 
\hdashline
succour& succor\\ 
\hdashline
arbour& arbor\\ 
\hdashline
catalogue& catalog\\ 
\hdashline
analog& analog\\ 
\hdashline
acknowledgement& acknowledgment\\ 
\hdashline
goitre& goiter\\ 
\hdashline
foetus& fetus\\ 
\hdashline
paediatrician& pediatrician\\ 
\hdashline
oesophagus& esophagus\\ 
\hdashline
manoeuvre& maneuver\\ 
\hdashline
oestrogen& estrogen\\ 
\hdashline
anaemia& anemia\\ 
\hline
\hline
\caption{Word list comprising British and American spelling variants.} 
\label{tab:spelling}
\end{longtable}

In our analysis, we consider two factors of differentiation between American and British English: spelling and vocabulary with different word lists used for each case. A given concept is expressed with two lexical alternations (either British or American) or two different spellings. The complete list of words and expressions employed in each case can be found, respectively, in Table~\ref{tab:vocabulary} (vocabulary) and Table~\ref{tab:spelling} (spelling). It is the result of compiling information in reference books~\cite{gramley03-1} and online sources such as the Oxford Dictionaries
(\url{https://en.oxforddictionaries.com/usage/british-and-american-terms}). In order to make sure that the items in the sources really represented British or American English, all the words in the list were subsequently checked in two widely used representative corpora of both varieties, namely, the British National Corpus (BYU-BNC) and the Corpus of Contemporary American English (COCA)~\cite{davies04-1, davies08-1}. Only pairs of words in which one of the members exhibits a significantly higher frequency in either of the two varieties were considered for inclusion in the list. Thus, for example, \textit{railway} is significantly more frequent in BYU-BNC whereas \textit{railroad} is significantly more frequent in COCA, which makes the pair valid for our purposes. Verbs ending in \textit{-ize}/-\textit{ise} (and the corresponding nouns in \textit{-isation} and \textit{-ization}) are, on the contrary, not considered because, as stated in Sec.~1, the spelling \textit{-ize} is common on both sides of the Atlantic, with the exception of \textit{analyse/analyze} and \textit{paralyse/paralyze}, which do form part of our list insofar as they have been found to be reliable British/American spellings. Inflectional forms (e.g., \emph{solicitor}, \emph{solicitors}, \emph{solicitor's}, \emph{solicitors'}) as well as derived (e.g., \emph{amphitheater}) and compound forms were also included in the search (e.g., \emph{sportscenter}). We are aware of the fact that departing from a list of Britishisms and Americanisms may appear to be a simplification of reality, because some Postcolonial Englishes may opt for vernacular forms, rather than for the British or the American one. However, our purpose is not to describe all varieties of English but to measure which of the two main inner-circle varieties is predominant in territories where English is used as a first, second and foreign language.

A word of caution is here needed. There exists certain semantic ambiguity in the selected lexical alternations. Nevertheless, this is an unavoidable effect that is inherent to computational studies on language variation (e.g., Refs.~\cite{goncalves14-1} and~\cite{huang16-1}). Our compromise is to keep the overall polysemy to a small degree while at the same time providing a selection of words sufficiently large to allow for a quantitative analysis. We have checked that the variants of each pair in our list can be exchanged, quite generally, in many contexts   and   are   thus   valid   for   the   aim of this work.

\subsection*{Metrics}

Language variation in space is analyzed by means of a grid of cells of $0.25^\circ\times0.25^\circ$ spanning the globe.
The polarization, $V_w^c$, for a concept $w$ in cell $c$ during the data collection period is defined as the ratio:
\begin{equation}
V_w^c=\frac{A_w^c-B_w^c}{A_w^c+B_w^c} ,
\end{equation}
where $A_w^c$ ($B_w^c$) is the number of American (British) forms of the concept $w$ observed in cell $c$. The polarization is then constrained to be in the $\left[-1,1\right]$ domain, with $-1$ corresponding to purely British and $1$ being purely American forms. 

The polarization of each cell, $V^c$, is then determined by taking the average polarization over all words observed in cell $c$:
\begin{equation}\label{eq_Vc}
V^c=\frac{\sum_w V_w^c}{W^c} ,
\end{equation}
where $W^c$ is the number of different words observed in cell $c$. Similarly, the polarization score of a country is defined as the average polarization taken over all the cells within that country. By considering the average polarization we are able to compare countries of varying sizes.

In the case of Twitter, the polarization signal is measured over the complete time period of the database since, as it is not long enough to allow for large variations in the language use patterns. On the other hand, when the time evolution of written language is considered with Google Books, space is not relevant, beyond the country of origin of the published book, and we add an index referring to the year $y$ considered. The polarization $V^y$ is then defined as:
\begin{equation}\label{eq_Vy}
V^y=\frac{\sum_w V_w^y}{W^y} \,,
\end{equation}
where $V_w^y$ is the concept polarization for year $y$ and
$W^y$ refers to all the books published in the country considered, the US or the UK, during year $y$.

\section*{Results}

\begin{figure}[!ht]
\begin{centering}
\includegraphics[width=0.95\columnwidth]{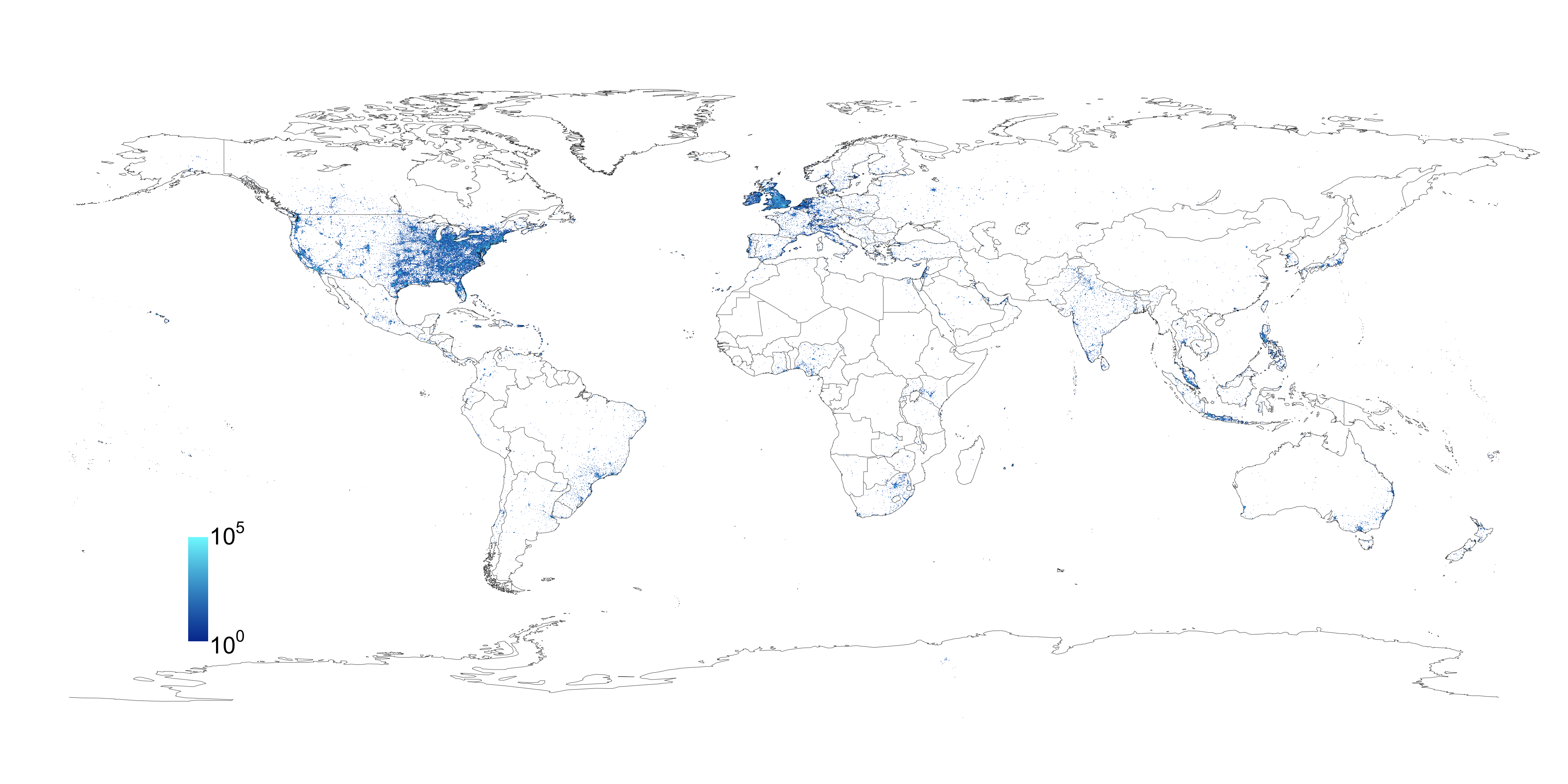}
\caption{\label{tweets} {\bf English tweets.} A heatmap showing the location of geolocated English tweets in our dataset that match our keywords.}
\end{centering}
\end{figure}

The analysis yields $30,898,072$ tweets matching the list of words. A heatmap illustrating the geographical distribution of matching tweets is shown in Fig~\ref{tweets}. The relation between bias of the data and population may lead to undesired fluctuations in our maps. This is particularly true in the US and the UK. On the other hand, in countries where English is not the mother tongue the real problem is the lack of data in certain cells. In the latter case, few English tweets may have a strong influence in the final value of the polarization for a given cell. We fix this issue in a twofold way. First, we impose for each cell a minimum threshold
of ten matches from our list of concepts. Second, we consider a sufficient number of cells. A balance between these quantities causes fluctuations to average out and they do not have a strong influence in the overall results.

\begin{figure}[!ht]
\begin{centering}
\includegraphics[width=0.95\columnwidth]{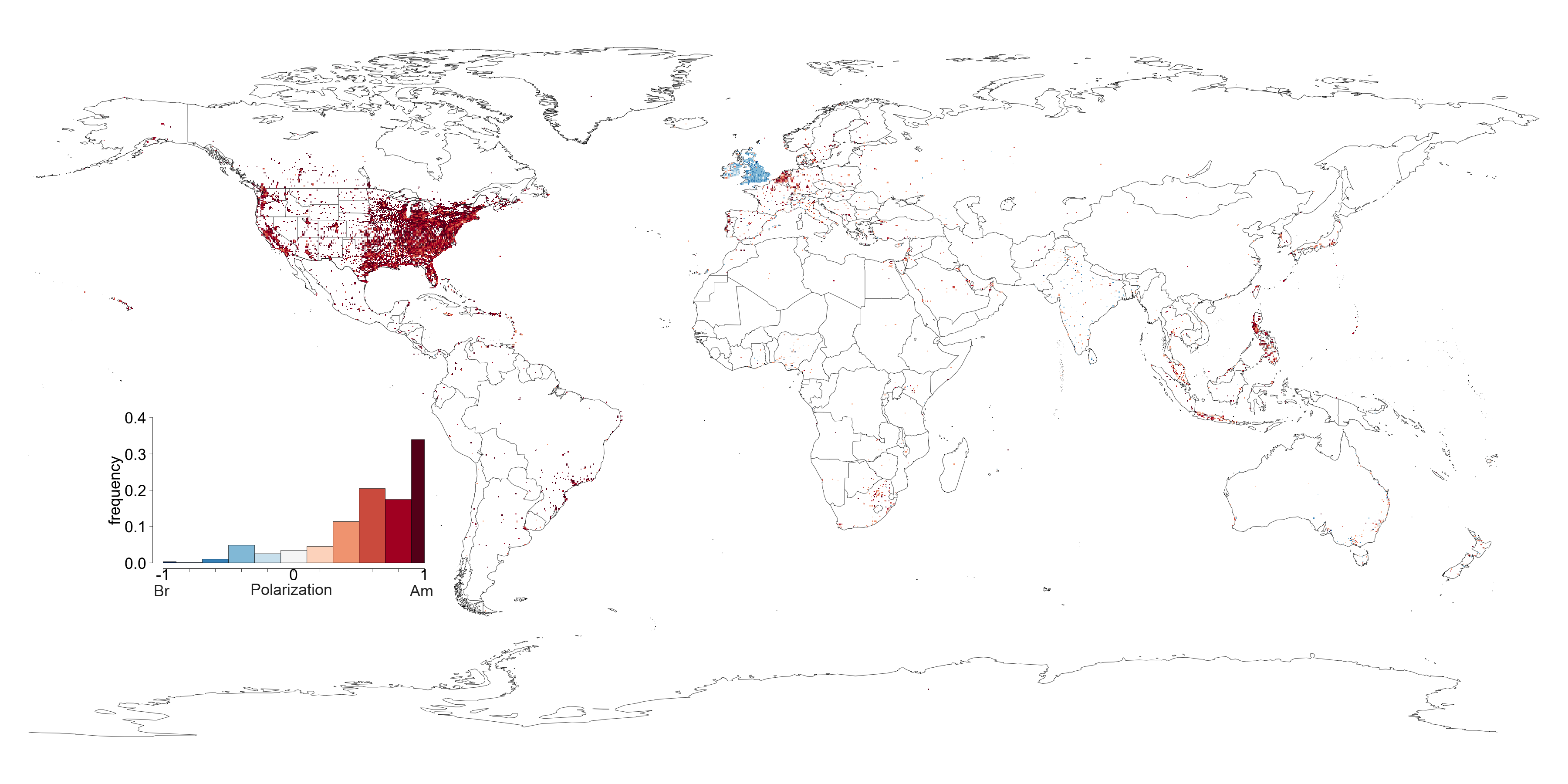}
\caption{\label{fig:vocabulary}{\bf Vocabulary.} The polarization ratio of each cell around the world according to the vocabulary used within each cell. The inset barplot is an histogram of the number of cells as a function of the ratio.}
\end{centering}
\end{figure}

\begin{figure}[!ht]
\begin{centering}
\includegraphics[width=0.45\columnwidth]{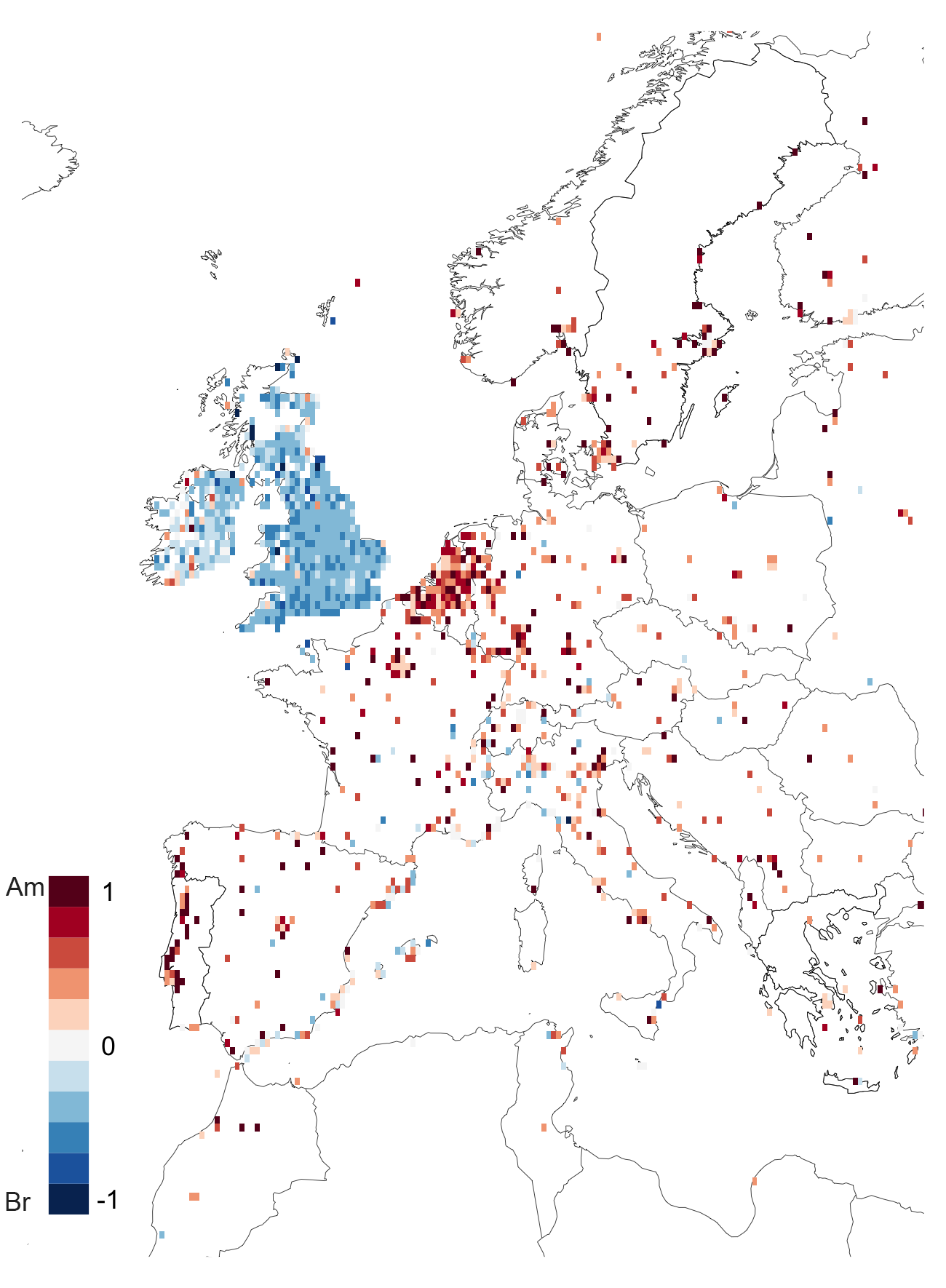}\includegraphics[width=0.45\columnwidth]{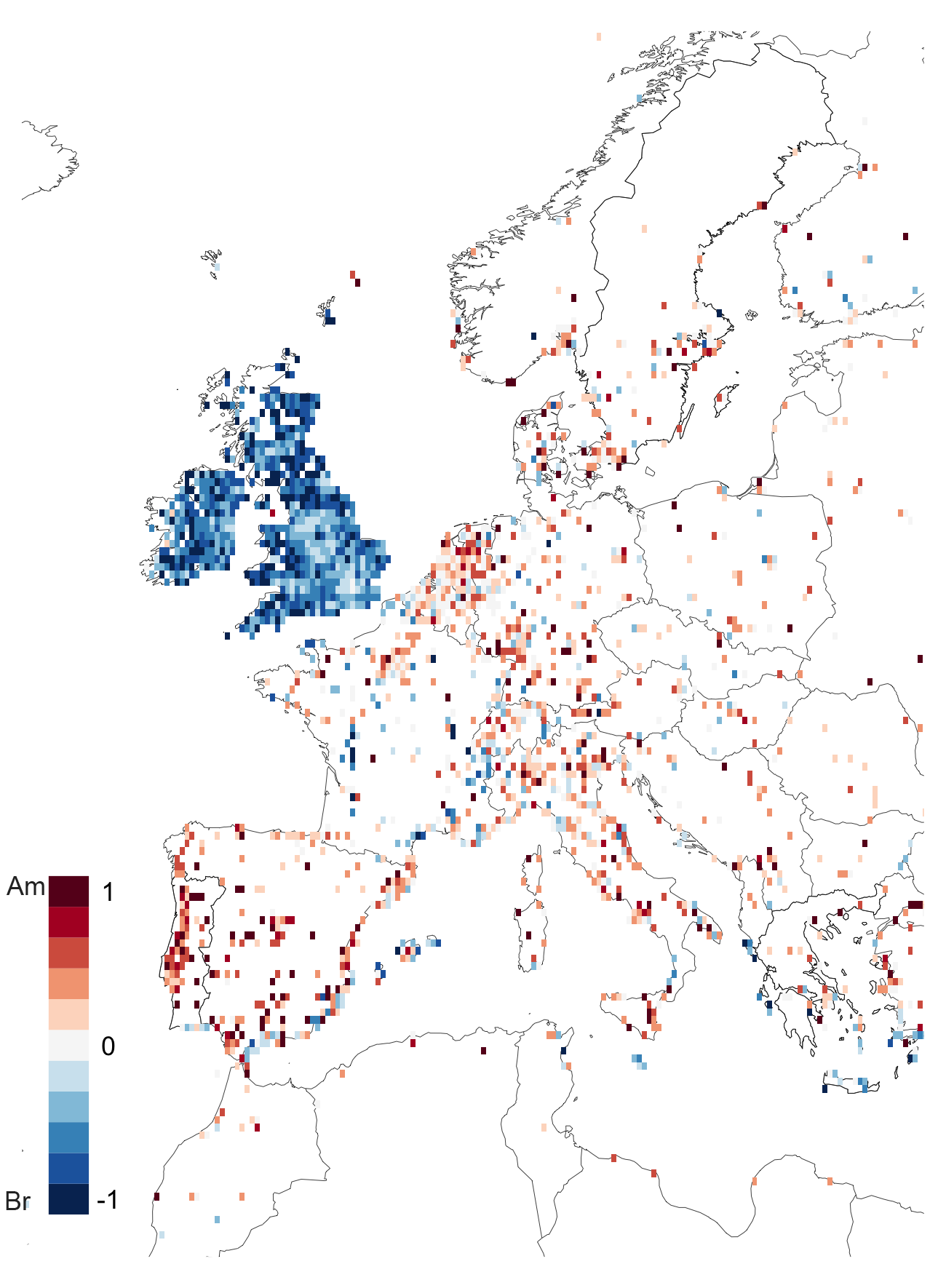}
\caption{\label{fig:Europe} {\bf Europe.} Side by side comparison of the vocabulary (left) and spelling (right) results for countries in continental Europe. The tension between British spelling and American vocabulary is clearly visible by the shift towards lighter shades of blue and darker shades of red between the left and the right plots.}
\end{centering}
\end{figure}

Let us start by considering how the vocabulary used for common terms such as \emph{lorry}/\emph{truck} or \emph{motorway}/\emph{freeway} changes around the world by defining the ratio of each cell as given by Eq.~\eqref{eq_Vc}. The results are plotted in Fig~\ref{fig:vocabulary}. Unsurprisingly, we find that the British Islands tend to be blue while the United States is predominantly red as befits the representatives of each trend. Interestingly, Western Europe where English teaching has traditionally followed British norms the American influence is undeniable. Most areas are depicted in various shades of red while some of the largest international metropolises such as Madrid, Paris, Amsterdam, Berlin, Milan or Rome are visible in light shades, indicating intermediate values, in no doubt due to their role as touristic and transportation hubs, see Fig~\ref{fig:Europe} (left). A more marked British influence is easily seen in former colonies such as South Africa, Australia, New Zealand (``the only large areas in the Southern hemisphere where English is spoken as a native language''~\cite{gramley03-1}), and which have reached a very advanced phase of development, according to Schneider's 2007 Dynamic Model~\cite{schneider07-1}) or India (where English is spoken as a non-native language, but which has followed an exonormative model, i.e., strongly based on British rules~\cite{schneider11-1}) displaying large areas of blue side by side with tell-tale patches of white in the most international areas such as Pretoria, Melbourne, Sidney, Auckland, New Delhi or Mumbai. Furthermore, countries such as the Philippines (one of the few postcolonial varieties of English with an American superstratum~\cite{schneider07-1}), as well as Taiwan, South Korea and Japan (where English is spoken as a second language) attest their strong American influence with full displays of red. 

\begin{figure}[!ht]
\begin{centering}
\includegraphics[width=0.95\columnwidth]{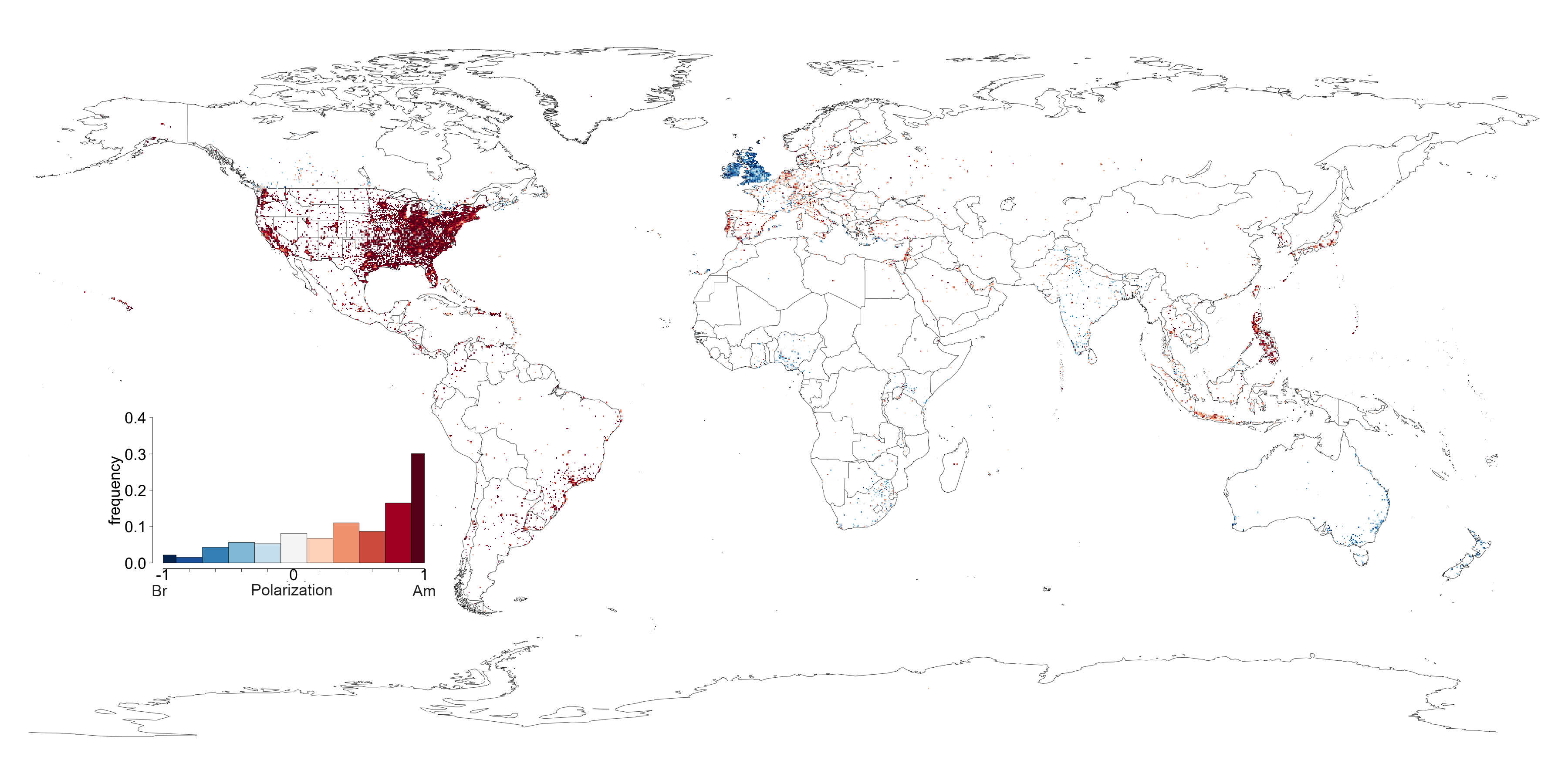}
\caption{\label{fig:spelling} {\bf Spelling.} The polarization of each cell around the world according to the spelling used within each cell. The inset barplot is an histogram of the number of cells as a function of the ratio observed.}
\end{centering}
\end{figure}

\begin{figure}[!ht]
\begin{centering}
\includegraphics[width=0.95\columnwidth]{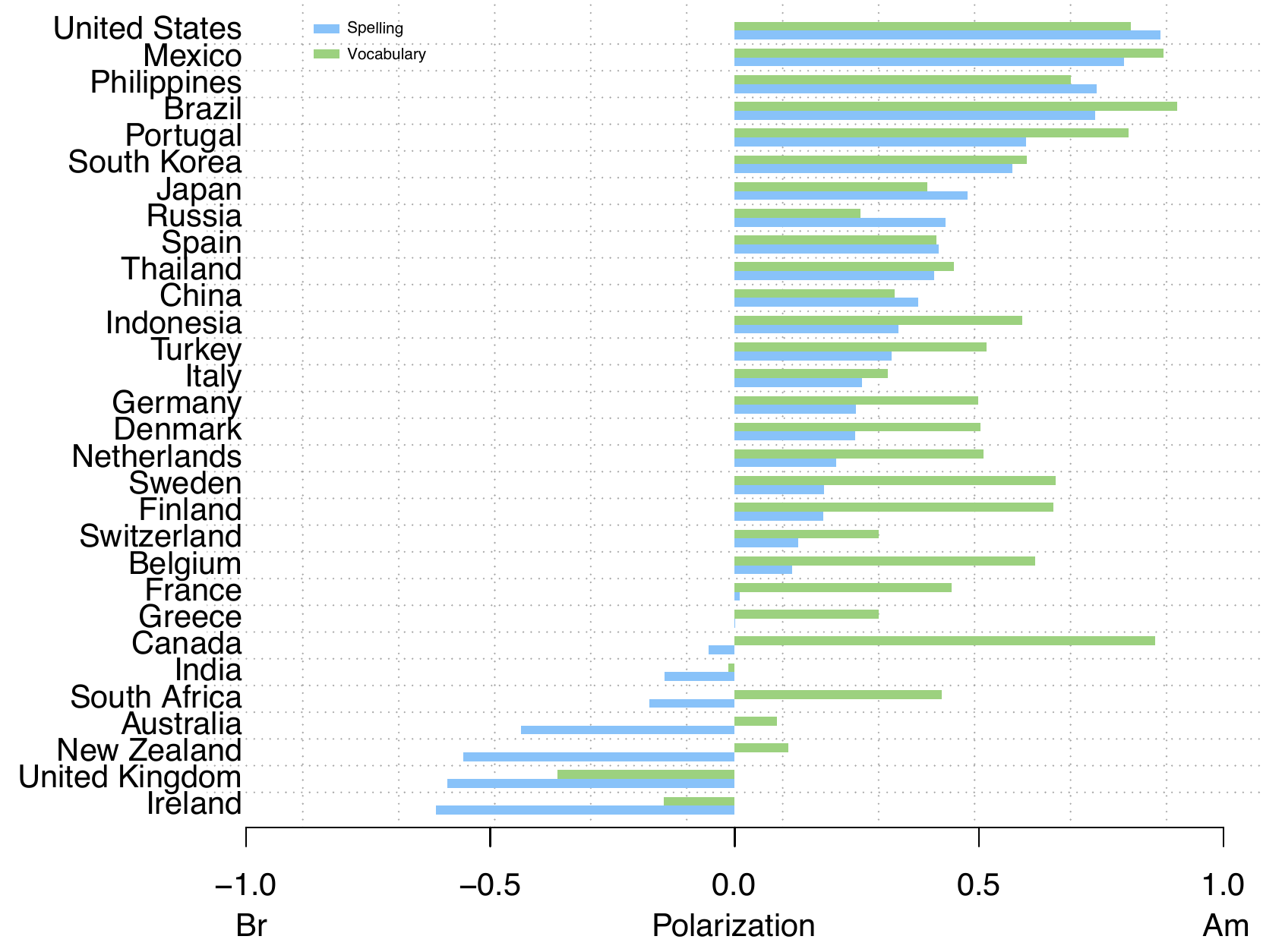}
\caption{\label{fig:countries} {\bf Countries.} Vocabulary and spelling polarization ratio by country.}
\end{centering}
\end{figure}

Regarding spelling, the case for American influence becomes even stronger as displayed in Fig~\ref{fig:spelling}. The British Isles attain significantly lighter shades of blue as do the former British colonies with South Africa, Australia and New Zealand becoming predominately red. This dichotomy between spelling and vocabulary, illustrated in Fig~\ref{fig:Europe} for Europe, is perhaps a testament to the conflicting forces of traditional formal education and media influence. Individuals who studied in school systems that subscribe to the British form of English are more prone to continue writing words in the way they originally learned them. However, through the influence of American dominated television and film industries they have acquired new (American) vocabulary. This can be clearly seen in Fig~\ref{fig:countries} where we plot the average polarization for both vocabulary and spelling for $30$ countries around the world,
including countries belonging to Kachru’s~\cite{kachru} Inner Circle, i.e., where English is spoken as a native language (e.g., UK, Ireland), Outer circle, i.e., where English is spoken as a second language (e.g., India, South Africa) and the expanding circle, i.e., where English is spoken as a foreign language (e.g., Portugal, Finland, Russia). Interestingly enough, in all expanding circle territories, American orthography and vocabulary dominate, and the same happens, obviously, in the United States and in the Philippines, a former American colony. The bottom part of the figure includes Inner and Outer circle varieties, where American vocabulary is also chosen over British forms,
with the notable exception of India, UK and Ireland, whose green bars are always towards the left hand (British) side of the ratio spectrum. India’s alignment with the UK is
clearly the result of an exonormative model and postcolonial prescriptivism in this former colony of the United Kingdom~\cite{schneider11-1,collins13-1}. Surprisingly,
we find that in some ex-colonies which still hold strong ties with 
the British empire, such as South Africa, Australia and New Zealand,
the drift towards American vocabulary is unmistakable.

\begin{figure}[!ht]
\begin{centering}
\includegraphics[width=0.95\columnwidth]{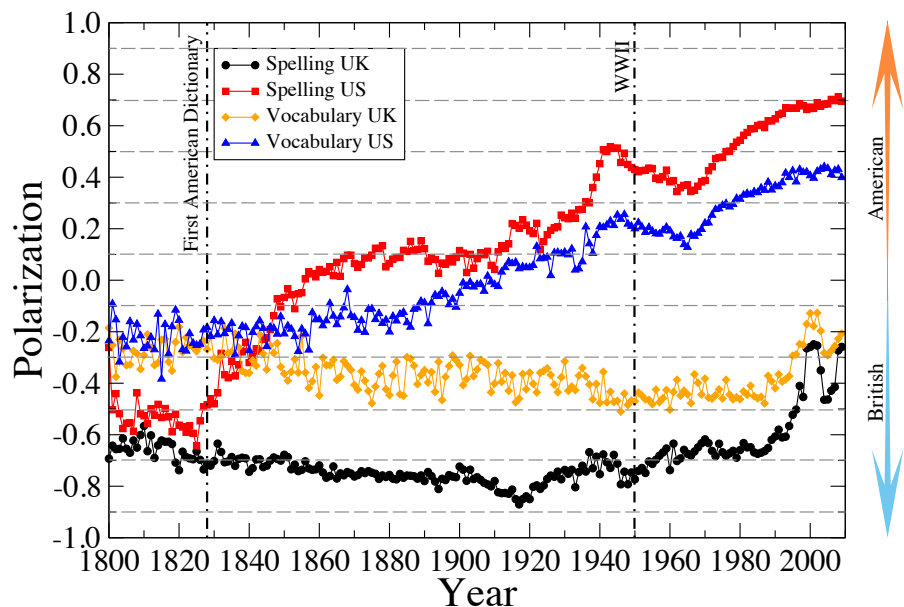}
\caption{\label{fig:books_final} {\bf Americanization of English over time.} Averaged polarization ratio of vocabulary and spelling for books published by US and UK publishing companies in the $1800-2010$ period.}
\end{centering}
\end{figure}

We now consider a temporal view of how English as a language is evolving. Using the word counts provided by the Google Books digitalization efforts, we measure the vocabulary and spelling average ratio per year [Eq.~\eqref{eq_Vy}]
for books published by American and British publishing houses. Considering the averages suffices for our purposes since averaging over the huge number of word instances in our two corpora ensures negligible error bars.
An analysis of the resulting timelines as shown in Fig~\ref{fig:books_final} provides several interesting insights. First, we can see that the divergence in spelling between the American and British forms has significantly increased in the last $200$ years. Indeed, from this time series we can pinpoint the beginning of the trend to around $1828$ when Noah Webster published {\em An American Dictionary of the English Language}~\cite{webster28-1} with the explicit goal of systematizing the way in which English was written in America. As~\cite{cassidy01-1} puts it: ``He is certainly responsible for establishing (though not inventing) the common differences between traditional British and American spellings'' the final \emph{-or} versus \emph{-our} in \emph{color}, \emph{labor}, \emph{savor}, and the like; \emph{-er} versus French \emph{-re} in \emph{theater}, \emph{center}, \emph{meter}; and the simplification of final \emph{-ck} as in \emph{physic}, \emph{music}, \emph{logic}. This is now considered to have been the first American English dictionary and it started the Merriam-Webster series of Dictionaries that is still dominant today. The US vocabulary curve follows a similar but less pronounced trend as it takes longer for new words to be created than for people to agree on a common spelling form.

Another interesting feature of these timelines is the pronounced ``Britishization'' of American English in the years following World War II as seen by the declining slope that extends until after $1960$. This can likely be explained by the large influx of European migrants that moved to America in search of a better life away from a destroyed or warring Europe. In the immediate aftermath of WWII, Congress passed the War Brides Act in $1946$ and the Displaced Persons Act in $1948$ to facilitate the immigration to the US by the people affected by the war. It is estimated that between $1941$ and $1950$ over $1$ Million people~\cite{abstract50}, mostly of European descent, immigrated to the United States that at the time had a population of $150$ million. In the following decade, this number doubled to over $2$ Million~\cite{abstract60}.

Interestingly, while the ratio timelines within the United Kingdom had been towards becoming ever more British, we find a significant change of trend in the last $20$ years of our dataset, corresponding to the period after the fall of the Berlin Wall and the end of the Cold War that left America as the world's only superpower. A position that was only reinforced with the advent and popularization of the Internet just a decade later. It is the status quo resulting from the aftermath of this trend that we are able to observe in the Twitter analysis above.

\section*{Conclusions}

The way in which languages evolve in time and change from place to place has long been the focus of much interest in the linguistic community. With the advent of new and extensive corpora derived from large scale online datasets we are now able to take on a more quantitative approach to tackling this fundamental question. In this work we analyze two datasets that, when taken together, are able to provide a bird's eye view of the way English usage has been changing over time and in different countries. 

The picture we are able to paint is particularly stark. The past two centuries have clearly resulted in a shift in vocabulary and spelling conventions from British to American. This trend is especially visible in the decades following WWII and the fall of the Berlin Wall. These historical events left the US as the only superpower and the influence it has exerted because of this on other cultures is evident at all levels. The presence of the American way of life, including cultural representations (literature, music, cinema and pop-culture products such as TV shows, computer games, etc.) can be felt all over the globe, as explained in Ref.~\cite{jones05} about the role played by the MTV as a powerful propagator of pop-culture. Naturally, the spread of the American culture is accompanied by the American linguistic variety, which ends up affecting (global) English, as we have shown, with some clearly identifiable exceptions.
Indeed, when we consider the current status quo as seen through the lens of Twitter, it becomes clear that only in the countries where British influence has been strongest, such as ex-colonies
with a strong exonormative influence (in Schneider’s terms~\cite{schneider07-1}),
are British conventions still dominant to some degree.

It should be noted that both datasets we utilize in our analysis are intrinsically biased. Books are typically written by cultural elites. Also, despite their increasing democratization, GPS enabled mobile devices are, in many countries, only available to middle and higher economic strata. As a result, there are certainly factors of linguistic evolution we are missing but the fact that both datasets agree on the general picture means that we are able to capture, at the very least, the underlying trends.

\section*{Acknowledgments} 

We acknowledge support from MINECO (Spanish Ministry of Economy, Industry and Competitiveness) (\url{http://www.mineco.gob.es/}), the Spanish Agency for Research AEI and the European Regional Development Funds (ERDF) under Grants No. FIS2015-63628-C2-2-R and FFI2017-82162. The funder had no role in study design, data collection and analysis, decision to publish, or preparation of the manuscript.

\section*{Supporting information}

This section contains an example of a python code used to query the Twitter streaming API. The idea is to collect only geolocated tweets enclosed in a given geographical area BOX limited by latitudes $y_0$ and $y_1$, and longitudes $x_0$ and $x_1$, the two corners of BOX being $(y_0,x_0)$ and $(y_1,x_1)$. This code is only for illustrative purposes. Twitter developers may change the format of the queries at any moment. The updated information on how to access the Twitter API can be found in the user guide at \url{https://developer.twitter.com/en/docs}

\begin{lstlisting}
from tweepy import Stream, OAuthHandler 

from tweepy.streaming import StreamListener 

CONSUMER_KEY = '' 
CONSUMER_SECRET = '' 
ACCESS_KEY = '' 
ACCESS_SECRET = '' 
BOX = [x0, y0, x1, y1] 

class MyStreamListener(StreamListener): 
def on_status(self, status): 
print(status) 

if __name__ == '__main__': 
auth = OAuthHandler(CONSUMER_KEY, CONSUMER_SECRET)
auth.set_access_token(ACCESS_KEY, ACCESS_SECRET) 

listen = MyStreamListener() 
stream = Stream(auth, listen, gzip=True) 
stream.filter(locations=BOX)
\end{lstlisting}

Regarding Google Books information, the data was downloaded from the American English and British English Ngram (N=1) viewer datasets available at \url{https://storage.googleapis.com/books/ngrams/books/datasetsv2.html}. These files enumerate how many times each word appears in the Google Books corpus each year.


\begin{thebibliography}{10}

\bibitem{labov63}
Labov W.
\newblock The social motivation of a sound change.
\newblock Word. 1963;19:273.

\bibitem{fishman67}
Fishman JA.
\newblock Bilingualism with and without diglossia; diglossia with a without
  bilingualism.
\newblock Word. 1967;XXIII:29.

\bibitem{crystal03-1}
Crystal D.
\newblock English as a Global Language.
\newblock Cambridge University Press; 2003.

\bibitem{jenkins13-1}
Jenkins J, Leung C.
\newblock English as a Lingua Franca.
\newblock The Companion to Language Assessment {I}{V}. 2013;13(95):1605.

\bibitem{mesthrie08-1}
Mesthrie R, Bhatt RM.
\newblock The Study of New Linguistic Varieties.
\newblock Cambridge University Press; 2008.

\bibitem{grieve16-1}
Grieve J.
\newblock Regional Variation in Written {A}merican {E}nglish.
\newblock Cambridge University Press; 2016.

\bibitem{pederson01-1}
Pederson L.
\newblock Dialects.
\newblock In: Algeo J, editor. The {C}ambridge History of the {E}nglish
  Language. Cambridge University Press; 2001. p. 253.

\bibitem{wolfram16-1}
Wolfram W, Shelling N.
\newblock American {E}nglish: Dialects and Variation.
\newblock Wiley-Blackwell; 2015.

\bibitem{leech09-1}
Leech G, Hundt M, Mair C, Smith N.
\newblock Change in Contemporary {E}nglish: A Grammatical Study.
\newblock Cambridge University Press; 2009.

\bibitem{baker17}
Baker P.
\newblock American and British English. Divided by a Common Language?
\newblock Cambridge University Press; 2017.

\bibitem{algeo01-1}
Algeo J.
\newblock External History.
\newblock In: Algeo J, editor. The {C}ambridge History of the {E}nglish
  Language. Cambridge University Press; 2001.

\bibitem{gramley03-1}
Gramley S, P{\"a}tzold KM.
\newblock Survey of Modern {E}nglish.
\newblock Routledge; 2003.

\bibitem{awonusi94-1}
Awonusi VO.
\newblock The {A}mericanization of {N}igerian {E}nglish.
\newblock World Englishes. 1994;13:75.

\bibitem{hansel13-1}
H{\"a}nsel EC, Deuber D.
\newblock Globalization, postcolonial {E}nglishes, and the {E}nglish language
  press in {K}enya, {S}ingapore, and {T}rinidad and {T}obago.
\newblock World Englishes. 2013;32:338.

\bibitem{davydova16-1}
Davydova J.
\newblock Indian {E}nglish quotatives in a real-time perspective.
\newblock In: Seoane E, Su{\'a}rez-G{\'o}mez C, editors. {W}orld {E}nglishes:
  {N}ew theoretical and methodological considerations. Benjamins; 2015. p. 173.

\bibitem{hackert15-1}
Hackert S.
\newblock Pseudotitles in {B}ahamian {E}nglish: A Case of {A}mericanization?
\newblock Journal of English Linguistics. 2015;43:143.

\bibitem{edwards16-1}
Hansen~Edwards JG.
\newblock Accent preferences and the use of {A}merican {E}nglish features in
  {H}ong {K}ong: a preliminary study.
\newblock Asian Englishes. 2016;18:197.

\bibitem{fuchs17-1}
Fuchs R.
\newblock The {A}mericanization of {P}hilippine {E}nglish: {R}ecent diachronic
  change in spelling and lexis.
\newblock Philippine ESL Journal. 2017;19.

\bibitem{mukherjee15-1}
Mukherjee J.
\newblock Response to {D}avies and {F}uchs.
\newblock English Worldwide. 2015;36:34.

\bibitem{venezky01-1}
Venezky RL.
\newblock Spelling.
\newblock In: Algeo J, editor. The {C}ambridge History of the {E}nglish
  Language. Cambridge University Press; 2001. p. 340.

\bibitem{nguyen15-1}
Nguyen D, Dogr\"{u}oz AS, Ros\'{e} CP, de~Jong F.
\newblock Computational sociolinguistics: {A} survey.
\newblock Computational linguistics. 2016;42:537.

\bibitem{melo16-1}
Melo F, Martins B.
\newblock Automated geocoding of textual documents: A survey of current
  approaches.
\newblock Transactions in GIS. 2016;21:3.

\bibitem{chambers98-1}
Chambers JK, Trudgill P.
\newblock Dialectology.
\newblock Cambridge University Press; 1998.

\bibitem{malmasi}
Malmasi S, Zampieri M, Ljubes\v{s}i\'{c} N, Nakov P, Ali A, Tiedemann J.
\newblock Discriminating between similar languages and {A}rabic dialect
  identification: A report on the Third {D}{S}{L} Shared Task.
\newblock Proceedings of the Third Workshop on NLP for Similar Languages,
  Varieties and Dialects (VarDial3). 2016; p. 1--14.

\bibitem{kulkarni16-1}
Kulkarni V, Perozzi B, Skiena S.
\newblock Freshman or Fresher? {Q}uantifying the Geographic Variation of
  Language in Online Social Media.
\newblock Proceedings of the Tenth International AAAI Conference on Web and
  Social Media. 2016; p. 613.

\bibitem{russ12-1}
Russ B.
\newblock Examining large-scale regional variation through online geotagged
  corpora.
\newblock ADS Annual Meeting. 2012;.

\bibitem{doyle14-1}
Doyle G.
\newblock Mapping Dialectal Variation by Querying Social Media.
\newblock In: Proceedings of the 14th Conference of the European Chapter of the
  Association for Computational Linguistics; 2014. p. 98--106.

\bibitem{goncalves14-1}
Gon\c{c}alves B, S\'{a}nchez D.
\newblock Crowdsourcing Dialect Characteriation Through {T}witter.
\newblock PLOS ONE. 2014;9:E112074.

\bibitem{goncalves16-1}
Gon\c{c}alves B, S\'{a}nchez D.
\newblock Learning about S{p}anish Dialects through {T}witter.
\newblock RILI. 2016;28:65--75.

\bibitem{donoso}
Donoso G, S\'{a}nchez D.
\newblock Dialectometric analysis of language variation in {T}witter.
\newblock Proceedings of the Fourth Workshop on NLP for Similar Languages,
  Varieties and Dialects (VarDial4). 2017; p. 16--25.

\bibitem{eisenstein14-1}
Eisenstein J, O'Connor B, Smith NA, Xing EP.
\newblock Diffusion of lexical change in social media.
\newblock PLOS ONE. 2014;9:E113114.

\bibitem{pavalanathan15-1}
Pavalanathan U, Eisenstein J.
\newblock Confounds and consequences in geotagged {T}witter data.
\newblock Proceedings of the Conference on Empirical Methods in Natural
  Language Processing. 2015; p. 2138–2148.

\bibitem{huang16-1}
Huang Y, Guo D, Kasakoff A, Grieve J.
\newblock Understanding {U}.{S}. regional linguistic variation with Twitter
  data analysis.
\newblock Computers, Environment and Urban Systems. 2016;54.

\bibitem{blodgett16-1}
Blodgett SL, Green L, O'Connor B.
\newblock Demographic Dialectal Variation in Social Media: A Case Study of
  {A}frican-{A}merican {E}nglish.
\newblock Proceedings of the 2016 Conference on Empirical Methods in Natural
  Language Processing. 2016; p. 1119–1130.

\bibitem{michel11-1}
Michel JB, Shen YK, Aiden AP, Veres A, Gray MK, Pickett JP, et~al.
\newblock {Quantitative Analysis of Culture Using Millions of Digitized Books}.
\newblock Science. 2011;331(6014):176--182.

\bibitem{pedersen12-1}
Pedersen AM, Tenenbaum JN, Havlin S, Stanley HE, Perc M.
\newblock Languages cool as they expand: Allometric scaling and the decreasing
  need for new words.
\newblock Sci Rep. 2012;2:943.

\bibitem{pechenik15-1}
Pechenick EA, Danforth CM, Dodds PS.
\newblock Is language evolution grinding to a halt? The scaling of lexical
  turbulence in English fiction suggests it is not.
\newblock Journal of Computational Science. 2017;21:24.

\bibitem{gerlach13-1}
Gerlach M, Altmann EG.
\newblock Stochastic Model for the Vocabulary Growth in Natural Languages.
\newblock Phys Rev X. 2016;3:021006.

\bibitem{mocanu13-1}
Mocanu D, Baronchelli A, Perra N, Gon{\c c}alves B, Vespignani A.
\newblock The {T}witter of {B}abel: Mapping World Languages through
  Microblogging Platforms.
\newblock PLOS One. 2013;8:E61981.

\bibitem{native1}
Lamanna F, Lenormand M, Salas-Olmedo MH, Romanillos G, Gon{\c c}alves B,
  Ramasco JJ.
\newblock Immigrant community integration in world cities.
\newblock PLOS ONE. 2016;13:e0191612.

\bibitem{native2}
Bassolas A, Lenormand M, Tugores A, Gon{\c c}alves B, Ramasco JJ.
\newblock Touristic site attractiveness seen through Twitter.
\newblock EPJ Data Science. 2016;5:12.

\bibitem{leetaru13}
Leetaru KH, Wang S, Cao G, Padmanabhan A, Shook E.
\newblock Mapping the global Twitter heartbeat: {T}he geography of Twitter.
\newblock First Monday. 2013;18:5.

\bibitem{pechenik15}
Pechinek EA, Danforth CM, Dodds PS.
\newblock Characterizing the {Google Books Corpus}: {S}trong Limits to
  Inferences of Socio-Cultural and Linguistic Evolution.
\newblock PLOS ONE. 2015;10:e0137041.

\bibitem{davies04-1}
Davies M.
\newblock {B}{Y}{U}-{B}{N}{C} (based on the {B}ritish {N}ational {C}orpus from
  {O}xford {U}niverstiy {P}ress); 2004.

\bibitem{davies08-1}
Davies M.
\newblock The Corpus of Contemporary {A}merican {E}nglish: 520 million words
  1990-present; 2008.

\bibitem{schneider07-1}
Schneider EW.
\newblock Postcolonial {E}nglish. Varieties around the World.
\newblock Cambridge University Press; 2007.

\bibitem{schneider11-1}
Schneider EW.
\newblock English around the World: An Introduction.
\newblock Cambridge University Press; 2011.

\bibitem{kachru}
Kachru BB.
\newblock Standards, codification and sociolinguistic realism: the {E}nglish
  language in the outer circle.
\newblock In: English in the world: Teaching and learning the language and
  literatures. Cambridge University Press; 1985. p. 11--30.

\bibitem{collins13-1}
Collins P.
\newblock Grammatical colloquialism and the English quasi-modals: a comparative
  study.
\newblock In: Mar{\'\i}n-Arrese JI, Carretero M, Hita JA, van~der Auwera J,
  editors. English modality: Core, Periphery and Evidentiality. Mouton de
  Gruyter; 2013.

\bibitem{webster28-1}
Webster N.
\newblock An American dictionary of the {E}nglish language.
\newblock {S. Converse}; 1828.

\bibitem{cassidy01-1}
Cassidy FG, Hall JH.
\newblock Americanisms.
\newblock In: Algeo J, editor. The Cambridge History of the English Language
  IV: English in North America. Cambridge University Press; 2001.

\bibitem{abstract50}
{Census Bureau}.
\newblock Statistical Abstract of the {U}.{S}. 1950; 1950.

\bibitem{abstract60}
{Census Bureau}.
\newblock Statistical Abstract of the {U}.{S}. 1960; 1960.

\bibitem{jones05}
Jones S.
\newblock {MTV}: The Medium was the Message.
\newblock Critical Studies in Media Communication. 2005;22:83.

\end{thebibliography}
\end{document}